\documentclass[10pt,twocolumn,letterpaper]{article}

\def \cD {\mathcal{D}}

\def \cL {\mathcal{L}}
\def \cI {\mathcal{I}}

\def \cQ {\mathcal{Q}}

\def \RR {\mathbb{R}}

\usepackage{iccv}
\usepackage{times}
\usepackage{epsfig}
\usepackage{graphicx}
\usepackage{amsmath}
\usepackage{amssymb}
\usepackage{multirow}
\usepackage{float}
\usepackage{subcaption}
\usepackage{mathtools}

\DeclarePairedDelimiter\floor{\lfloor}{\rfloor}

\usepackage[pagebackref=true,breaklinks=true,letterpaper=true,colorlinks,bookmarks=false]{hyperref}

\iccvfinalcopy 

\ificcvfinal\fi

\begin{document}

\title{AdvDrop: Adversarial Attack to DNNs by Dropping Information}

\author{Ranjie Duan\textsuperscript{1,}\textsuperscript{2}\footnotemark[1]\ \  
Yuefeng Chen\textsuperscript{2}\ \
Dantong Niu\textsuperscript{3}\ \
Yun Yang\textsuperscript{1} \footnotemark[2] \ \ 
A. K. Qin\textsuperscript{1} \footnotemark[2] \ \
Yuan He\textsuperscript{2}\\
\textsuperscript{1}Swinburne University of Technology, Australia \ \ 
\textsuperscript{2}Alibaba Group, China\\
\textsuperscript{3}University of California, Berkeley, USA \ \  
}
\maketitle

\renewcommand{\thefootnote}{\fnsymbol{footnote}} 
\footnotetext[1]{Work done when Ranjie Duan interns at Alibaba Group, China} 
\footnotetext[2]{Correspondence to: A. K. Qin \& Yun Yang} 
\footnote{Code is available at https://github.com/RjDuan/AdvDrop}

\begin{abstract}
Human can easily recognize visual objects with lost information: even losing most details  with only contour reserved, e.g. cartoon. However, in terms of visual perception of Deep Neural Networks (DNNs), the ability for recognizing abstract objects (visual objects with lost information) is still a challenge. In this work, we investigate this issue from an adversarial viewpoint: will the performance of DNNs decrease even for the images only losing a little information? 
Towards this end, we propose a novel adversarial attack, named \textit{AdvDrop}, which crafts adversarial examples by dropping existing information of images.
Previously, most adversarial attacks add extra disturbing information on clean images explicitly. Opposite to previous works, our proposed work explores the adversarial robustness of DNN models in a novel perspective by dropping imperceptible details to craft adversarial examples. We demonstrate the effectiveness of \textit{AdvDrop} by extensive experiments, and show that this new type of adversarial examples is more difficult to be defended by current defense systems. 
\end{abstract}

\section{Introduction}
Deep Neural Networks (DNNs) have demonstrated their outstanding performance across many applications such as computer vision \cite{he2016deep} and natural language processing \cite{zeng2019dirichlet}. Though DNNs have great achievement in these tasks, especially in computer vision, they are known to be vulnerable to adversarial examples. Adversarial examples of DNNs were first discovered by Szegedy et al. \cite{szegedy2013intriguing}, which are crafted by adding malicious perturbation on clean images to generate undesirable consequences. Various methods have been proposed to generate adversarial examples \cite{goodfellow2014explaining, madry2017towards,carlini2017towards}. Typically, the generated adversarial perturbation is bounded by a small norm ball, which guarantees the resultant images ``look like" benign images.
\begin{figure}[ht]
 \begin{center}
    \includegraphics[width = \linewidth]{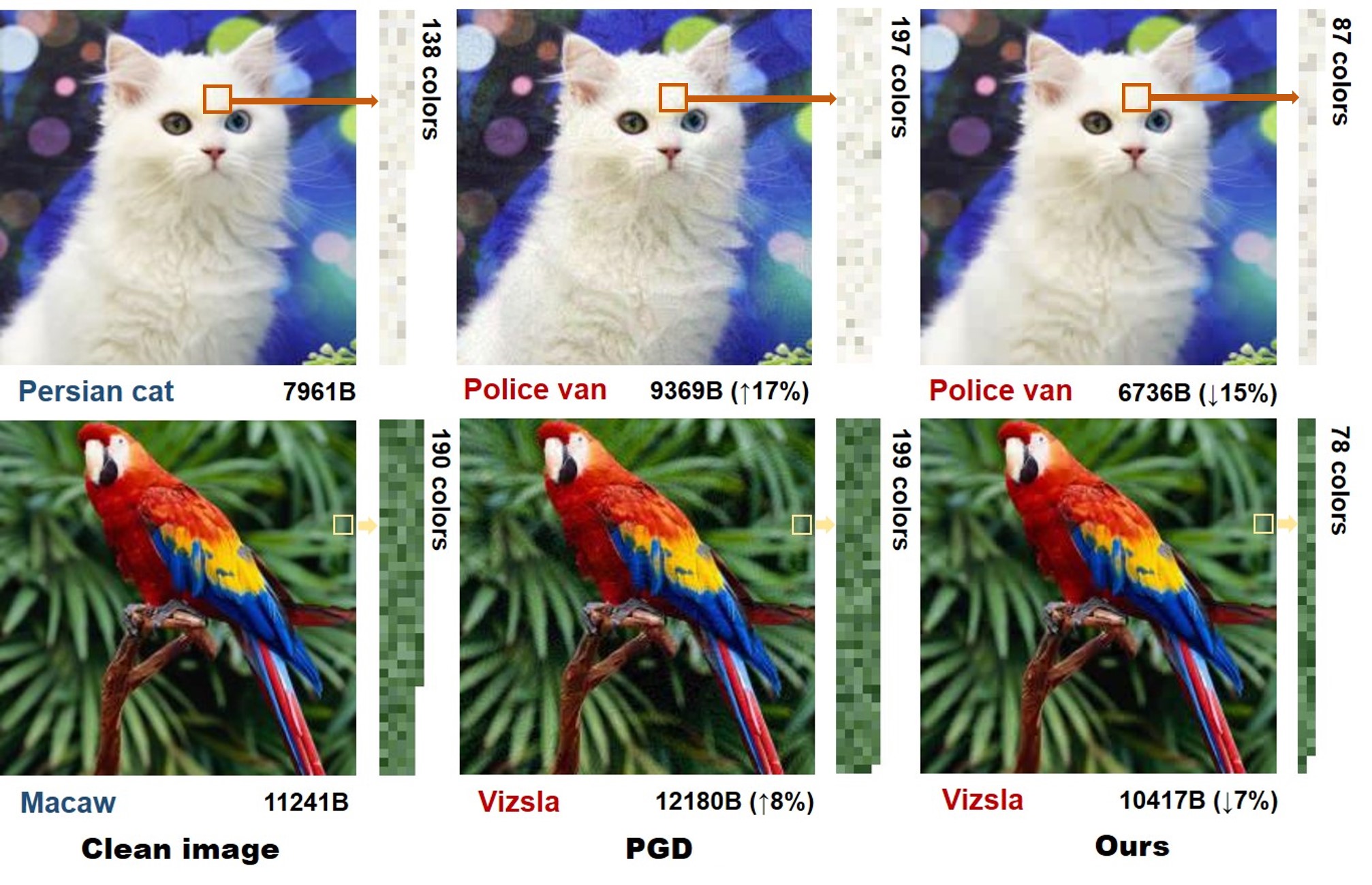}
  \end{center}
  \caption{\textbf{Adv. images generated by PGD and \emph{AdvDrop}.} Compared to the clean images, the adversarial images generated by \emph{AdvDrop} have fewer details composed of fewer colors, with the decreasing in size (by 15\% and 7\%). }
  \label{fig:exp}
\end{figure}

Interestingly, Ilyas et al. \cite{ilyas2019adversarial} empirically demonstrated that adversarial perturbation can be non-robust features for DNNs. That is to say, regarding adversarial perturbation, they are meaningful features for DNNs, but meaningless and imperceptible for humans.
So we wonder, whether it is possible to craft adversarial examples in an opposite paradigm? Rather crafting adversarial examples by adding adversarial perturbation (or non-robust features) on clean images, we drop certain features from clean images that are imperceptible to humans but essential for DNNs which further lead to DNNs failing to recognize the resultant images.
\begin{figure*}[ht]
 \begin{center}
    \includegraphics[width = 0.8\linewidth]{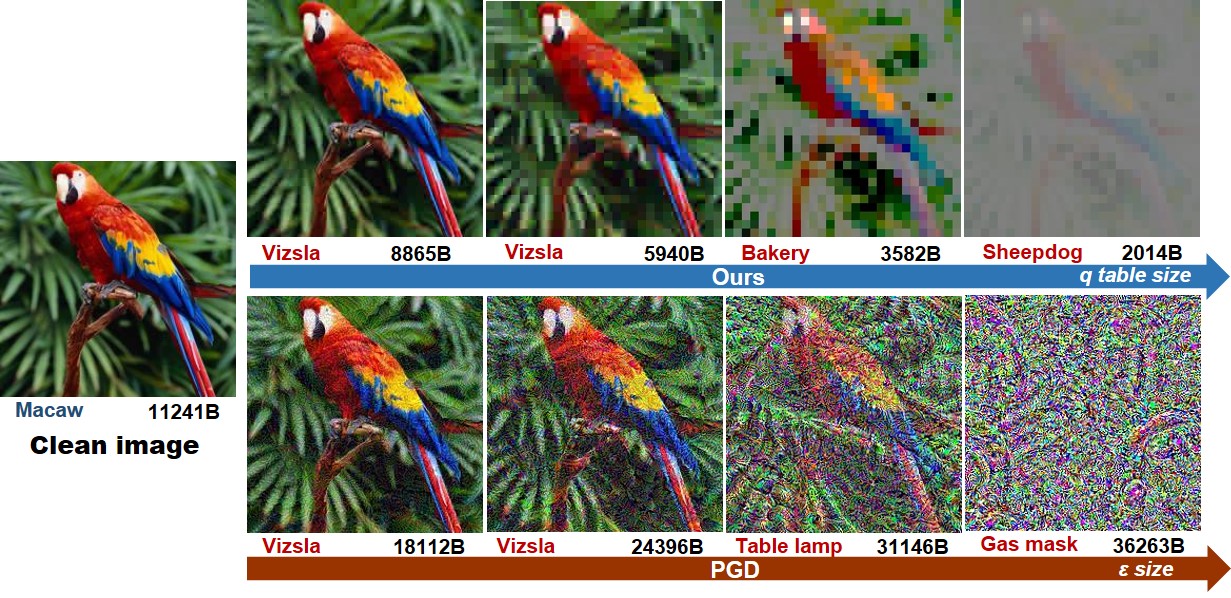}
  \end{center}
  \caption{\textbf{Interpolation between the clean image and adversarial images generated by \emph{AdvDrop} and PGD.}}
  \label{fig:interpolation}
\end{figure*}

Towards this end, we propose a novel adversarial attack named \emph{AdvDrop}, which crafts adversarial images by dropping less perceptible details from clean images. For example, as shown in Figure \ref{fig:exp},  both adversarial images generated by PGD \cite{madry2017towards} and \emph{AdvDrop} look indistinguishable from the clean images at first glance. However, when you look closely, PGD generates \textbf{extra} details (composed of more colors) at the cost of extra storage (larger image size). In contrast, the proposed $AdvDrop$ drops \textbf{existing} details such as subtle texture-like information from clean images, and the local patch is composed of less colors compared with the other images. As the figure indicates, the lost brittle details from benign images result in DNNs failing to recognize the resultant images correctly.

Dropping information of images can be achieved in either spatial domain (\eg color quantization \cite{heckbert1982color, orchard1991color}) or frequency domain (\eg JPEG compression \cite{wallace1992jpeg}). In our work, we consider developing proposed $AdvDrop$ in frequency domain. Principally, we can drop various features of an image to generate adversaries. This preliminary study is focused on the frequency domain because we choose to use ``image details" as the feature of interest to be dropped, which can be well quantized in the frequency domain. This choice is motivated by native insensitivity of human eyes to fine image details. In this work, $AdvDrop$ first transforms images from spatial domain to frequency domain, then reduces some frequency components of the transformed images quantitatively. Figure \ref{fig:interpolation} shows the process of $AdvDrop$, which performs attack following an opposite mechanism to PGD. The proposed \emph{AdvDrop} starts from removing subtle details (\eg textures) and the resultant image is almost indistinguishable from the clean one. When increasing the amount of dropped information, the resultant adversarial image finally turns to be somewhat ``blank". Here,``blank" denotes pure color which presents (almost) no information for recognizing a specific object for DNNs. 

We then perform comprehensive evaluation on the proposed $AdvDrop$. It can achieve high attack success rates in both targeted and untargeted settings on ImageNet \cite{deng2009imagenet}. We also evaluate the effectiveness of $AdvDrop$ in terms of defense methods. Various defense methods have been proposed to defend against adversarial examples \cite{madry2017towards, athalye2018obfuscated, prakash2018deflecting, xu2017feature}. Current defense methods are less effective against adversarial examples generated by \emph{AdvDrop} as they are generated with a rather different paradigm. Moreover, since the adversaries are generated by \emph{AdvDrop} via losing information, they are somewhat robust to denoising-based defenses.
\begin{figure}[hbt]
 \begin{center}
    \includegraphics[width = \linewidth]{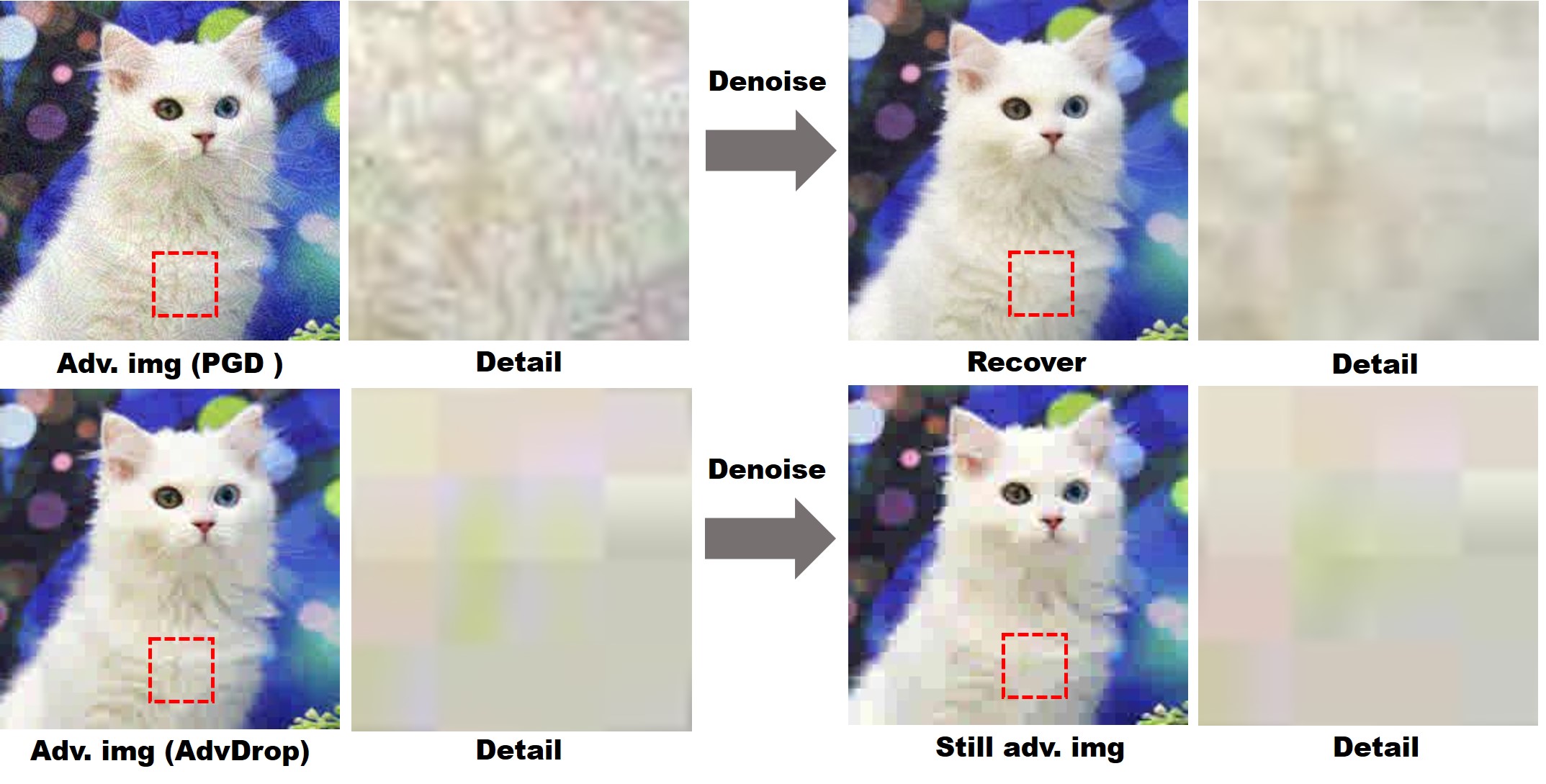}
  \end{center}
  \caption{\textbf{Adversarial images under denoising-based defense.} Adversarial perturbation generated by PGD could be mitigated by applying denoising strategies, but with almost no effect on adversaries generated by \emph{AdvDrop}. }
  \label{fig:defense}
\end{figure}
Typically, the denoising-based method removes the generated adversarial perturbation and accordingly defends against adversaries (Figure \ref{fig:defense}). For adversaries generated by $AdvDrop$, however, denoising-based defenses take no effect and the resultant images are still adversarial for DNNs. We hope this finding will motivate devising more effective defense approaches against \emph{AdvDrop}. In addition, to better understand the mechanism of \emph{AdvDrop} and the properties of generated adversarial examples by \emph{AdvDrop}, we provide visualizations of the dropped information by \emph{AdvDrop}, and perform a further analysis together with the attention of the DNNs.
In summary, this paper has made the following contributions:
\begin{itemize}
    \item {We propose a novel adversarial attack named $AdvDrop$, which is a totally different paradigm from previous attacks.  \emph{AdvDrop} crafts adversarial images by dropping \textbf{existing} details of clean images. It opens new doors to generate adversarial attacks for DNNs. }
    \item We conduct comprehensive experiments and demonstrate the effectiveness of $AdvDrop$ on targeted and untargeted attack settings. We also empirically show that current defense methods become less effective against adversarial examples generated by $AdvDrop$ compared to other attacks. 
    \item Finally, we visualize the dropped information and the attention of the DNNs to interpret the adversaries generated by $AdvDrop$.
\end{itemize}
This paper is organized as follows. Background and related work are discussed in Section \ref{sec:bg}. Our proposed approach is described in Section \ref{sec:approach}, and evaluated in Section \ref{sec:experiments}. Section \ref{sec:conclusion} concludes the paper and points out future work.

\section{Background and Related Work}\label{sec:bg}
\subsection{Adversarial attacks and defenses}
Adversarial attack was first proposed by Szegedy et al. \cite{szegedy2013intriguing}, aiming to generate perturbation superimposed on clean images to fool a target model. Adversarial attacks can be either digital-setting \cite{goodfellow2014explaining,kurakin2016adversarial, madry2017towards} or physical-setting \cite{eykholt2018robust,sharif2016accessorize,duan2020adversarial}, where most attacks are developed in digital-setting.
Specifically, given a target model $f$, adversarial example $x'$ can be crafted by either following the direction of adversarial gradients \cite{goodfellow2014explaining,kurakin2016adversarial, madry2017towards} or optimizing perturbation with a given loss \cite{carlini2017adversarial,chen2018ead}. For most adversarial attacks, the generated adversarial perturbation is bounded by a $l_p$ norm ball. Some works \cite{guo2020low, sharma2019effectiveness} propose generating adversarial examples in the frequency domain for either improving the efficiency in black-box setting or transferability. Roughly speaking, these adversarial attacks can be somewhat formulated as $x'=x+\delta$, where $\delta$ represents additive adversarial perturbation. Conversely, the proposed $AdvDrop$ crafts adversarial examples with an opposite paradigm $x'=x-\delta$. Note here `+' and `-' are not simple ``\textbf{add}" or ``\textbf{subtract}" operations over the values of $x$, but denote whether $\delta$ is extra information created by attacks or existing information of clean images dropped by $AdvDrop$. 

Other works also explore adversarial examples by making modifications or replacement on the secondary attributes (\eg color, lighting \cite{hosseini2018semantic,shamsabadi2020colorfool, zhao2020towards, zeng2019adversarial,liu2018beyond,zhao2018generating, duan2021adversarial}, texture \cite{wiyatno2019physical,duan2020adversarial}) to generate adversarial examples. We also note some other works \cite{guo2020watch, dabouei2020smoothfool} proposed attack based on motion blurring or smoothing, which also lose details of the clean image after the attack. However, their purposes are different from ours. This work aims to demonstrate the effectiveness of the proposed mechanism.

Many adversarial defense techniques have been proposed. Madry et al. \cite{madry2017towards} proposed adversarial training, which is arguably one of the most effective defense against adversarial attacks. Adversarial training is a data augmentation technique that trains DNNs on adversarial examples rather than natural examples. However, adversarial training is both time and computation consuming, due to the generation of adversarial examples and extra training epochs to fit adversarial examples. There are also preprocessing based methods, which process the inputs with certain transformations to remove the adversarial noise, and then send these inputs to the target model \cite{dziugaite2016study, das2017keeping, das2018shield, guo2018countering, prakash2018deflecting}. We consider both types of defenses during the evaluation to evaluate the effectiveness of \emph{AdvDrop}.

\subsection{Image compression methods}
Image compression methods fall into two categories, lossless compression, \eg, PNG \cite{boutell1997png}, and lossy compression, \eg, JPEG \cite{wallace1992jpeg,skodras2001jpeg}. 
JPEG compression applies discrete cosine transform (DCT) on patches which transforms images from spatial domain to frequency domain. DCT works by separating the image into different parts of different frequencies. Then JPEG applies quantization matrix (designed based on human vision) on transformed images dropping most of high-frequency components. In detail, higher frequency components of transformed images are rounded to zero, and finally reduce the size of original images.
Recently, deep learning based methods have been investigated for image compression problems. Both  CNN-based methods \cite{li2018learning, agustsson2019generative, agustsson2017soft} and RNN-based methods \cite{toderici2017full, johnston2018improved} are investigated. However, deep learning based compression methods are time-consuming and require pre-training.
Due to both efficiency and convenience concerns, we follow the design of JPEG to develop our proposed \emph{AdvDrop}. 
\begin{figure*}[ht]
 \begin{center}
    \includegraphics[width = \linewidth]{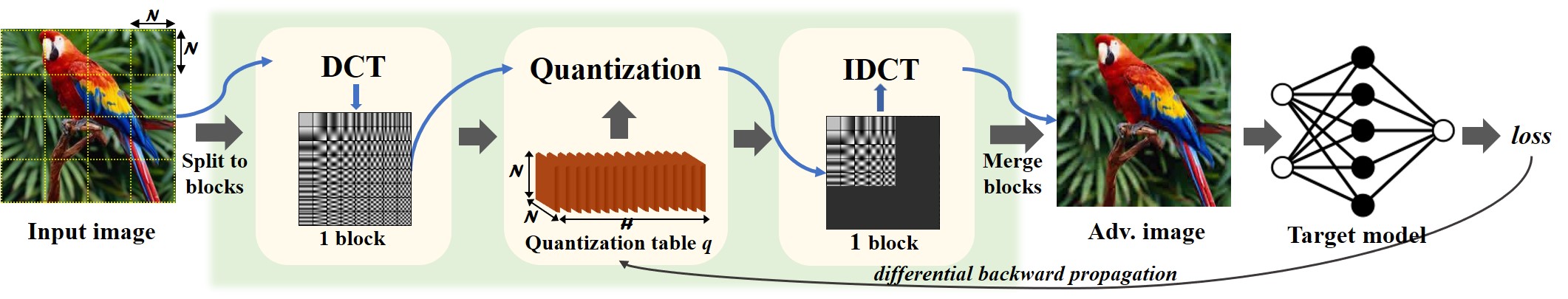}
  \end{center}
  \caption{\textbf{Pipeline.}}
  \label{fig:process}

\end{figure*}

\section{Approach} \label{sec:approach} 
\subsection{Overview}
Given a clean image $x \in \RR^m$ with class label $y$, a DNN classifier $f: \RR^m \to \{1, \cdots, k\}$ which maps image pixels to a discrete label set, and a target class $y_{adv} \neq y$ for targeted attack. The goal of adversarial attack is to find an adversarial example $x'$ for clean image $x$ by solving the optimization problem $\cL_{adv}(\cdot)$, which is the adversarial loss leading to $f(x') \neq y $ or $f(x') = y_{adv}$. Typically, $x'$ is restricted by $l_{\infty}$ norm ball: $\|x' - x  \|_{\infty} < \epsilon$. Our goal is to develop a mechanism that drops information from benign images to craft adversarial images. \emph{AdvDrop} is composed of several parts: 
\begin{itemize}
    \item \textbf{Adversarial loss: }The proposed method optimizes over quantization table \textit{q} by minimizing adversarial loss $\cL_{adv}{(\cdot)}$.
    \item \textbf{Discrete Cosine Transform (DCT): }DCT transforms the input image $x$ from spatial domain to frequency domain. DCT is denoted as $\cD{(\cdot)}$ in the following.
    \item \textbf{Inverse Discrete Cosine Transform (IDCT): } IDCT transforms image's signals from frequency back to spatial domain, denoted as $\cD_{\cI}(\cdot)$. 
    \item \textbf{Quantization: }Quantization serves as the core process to drop information by applying quantization table \textit{q}, which is optimized during the attack. We denote common quantization as $\cQ{(\cdot)}$. However, in our work, we adopt a differential quantization process denoted as $\cQ_{diff}(\cdot)$. Note in the following, either $\cQ(\cdot)$ or $\cQ_{diff}(\cdot)$ represents a complete quantization-dequantization process.
\end{itemize}
In summary, our proposed $AdvDrop$ first transforms clean images from spatial to frequency domain, then applies quantization to drop some specific frequencies of the transformed image, followed by inverting the frequency signals of images back to spatial domain. During optimization, $AdvDrop$ only tunes the value of quantization table \textit{q} bounded by $\epsilon$. Formally, we denote our final objective as:
\begin{equation}\label{eq:obj}
\begin{aligned}
 \min\limits_{q} \:\: \cL_{adv}(x', y),&\:\: \textrm{where} \:x'= \cD_{\cI}(\cQ_{diff}{(\cD(x), \textit{q})}) \\
\textrm{s.t. }\: &\left\|q - q_{init}\right\|_{\infty} < \epsilon
\end{aligned}
\end{equation}
where $q_{init}$ is the initial value of quantization table \textit{q}. We set $q_{init} = \textbf{1}$. We increase the value of quantization table \textit{q} gradually to drop the information of given image during the optimization. $\cQ_{diff}$ represents a differential quantization function, which allows \emph{AdvDrop} to compute the gradients during the backward propagation.
The overview of $AdvDrop$ is illustrated in Figure \ref{fig:process}. 
\subsection{Adversarial loss}
For adversarial loss $\cL_{adv}(\cdot)$, we use the following cross-entropy loss:
\begin{equation}\label{eq:adv_loss}
\cL_{adv}= 
  \begin{cases}
    \log(p_{y}(x')), &\text{for untargeted attack},\\
    -\log(p_{y_{adv}}(x')) , &\text{for targeted attack}\\
    \end{cases}
\end{equation}
where $p(\cdot)$ is the probability output (softmax on logits) of the target model $f$ with respect to class $y_{adv}$ or $y$. By minimizing loss $\cL_{adv}$, $AdvDrop$ optimizes the quantization table \textit{q} to selectively drop the information of input image $x$, in order to mislead the target model $f$ finally. 
\subsection{Transformation}
We introduce both DCT and IDCT in this part. DCT serves as a transformation of images from spatial domain to frequency domain, which expresses a finite sequence of data points in terms of a sum cosine functions oscillating at different frequencies. Note that there are also other methods enabling transformation of images from spatial to frequency domain, such as discrete Fourier transform etc. We focus on DCT as it enables the proposed $AdvDrop$ to have more flexibility in selecting what and where information to drop. 
 
Before applying DCT, we first split the original images into blocks with size $N\times N$ as shown in Figure \ref{fig:process}. Our proposed method splits images into patches with size $N = 8$ \cite{skodras2001jpeg, wallace1992jpeg} for all the experiments by mainly considering computation cost and perceptual quality. As the correlation between pixels within the patch decreases with increasing in patch size, the quantization results in larger distortion.
For each block, the value of pixels are adjusted to be symmetric with respect to zero. The mathmatical definition of DCT is: \\
\begin{equation}\label{eq:dct}
\begin{aligned}
    \cD{(x)}_{[u,v]} = &\frac{1}{\sqrt{2N}}C(u)C(v)\sum_{x=0}^{N-1}\sum_{y=0}^{N-1}x[k,m]\\
    &cos[\frac{(2k+1)i\pi}{2N}]cos[\frac{(2m+1)j\pi}{2N}],
\end{aligned}
\end{equation} 
In which, Eq. \ref{eq:dct} computes the $u,v^{th}$ entry of $\cD{(x)}$. $x[k, m]$ with the value on coordinate $(k, m)$ of image $x$. $N$ is the size of the block. A concrete example can be seen in Figure \ref{fig:process}, where $\cD$ transforms input image from spatial domain to a series of blocks in frequency domain. 

IDCT is the inverse process of DCT, which serves as recovering the signals of input image from frequency back to spatial domain. Due to the page limit, more details can be referred to \cite{ahmed1974discrete}. Note that either DCT or IDCT is lossless. The information is only lost during the quantization. We then discuss how the quantization is capable of dropping information. 

\subsection{Quantization}
The quantization is done with two operations: rounding and truncation. The former maps the original value to its nearest quantization point, while the latter confines the range of quantized values. A common complete quantization-dequantization process $\cQ(\cdot)$ is defined by:

\begin{equation}\label{eq:round}
\cQ(x, \Delta) = \floor{\frac{x+0.5}{\Delta}}\cdot\Delta,\:\: 
that\:\: \cQ(x, \Delta) \in [\epsilon_{min}, \epsilon_{max}]
\end{equation}
where $\Delta$ denotes the interval length, which decides the nearest quantization point for value of $x$, serves as a quantizer. Intuitively, the larger of $\Delta$, the smaller of the length of the set of quantized values after quantization. The quantized values are constrained in a valid range $[\epsilon_{min}, \epsilon_{max}]$.  

\begin{figure}[htb]
 \begin{center}
    \includegraphics[width = \linewidth]{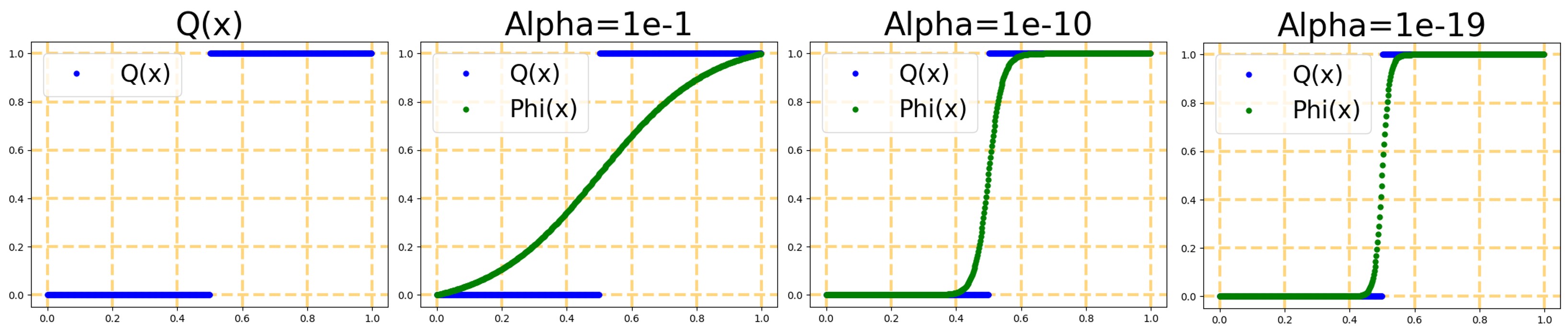}
  \end{center}

  \caption{\textbf{Illustration of the usage of $\varphi(x)$ and $\alpha$.}  }
  \label{fig:phi-x}

\end{figure}

\begin{figure}[htb]
 \begin{center}
    \includegraphics[width = \linewidth]{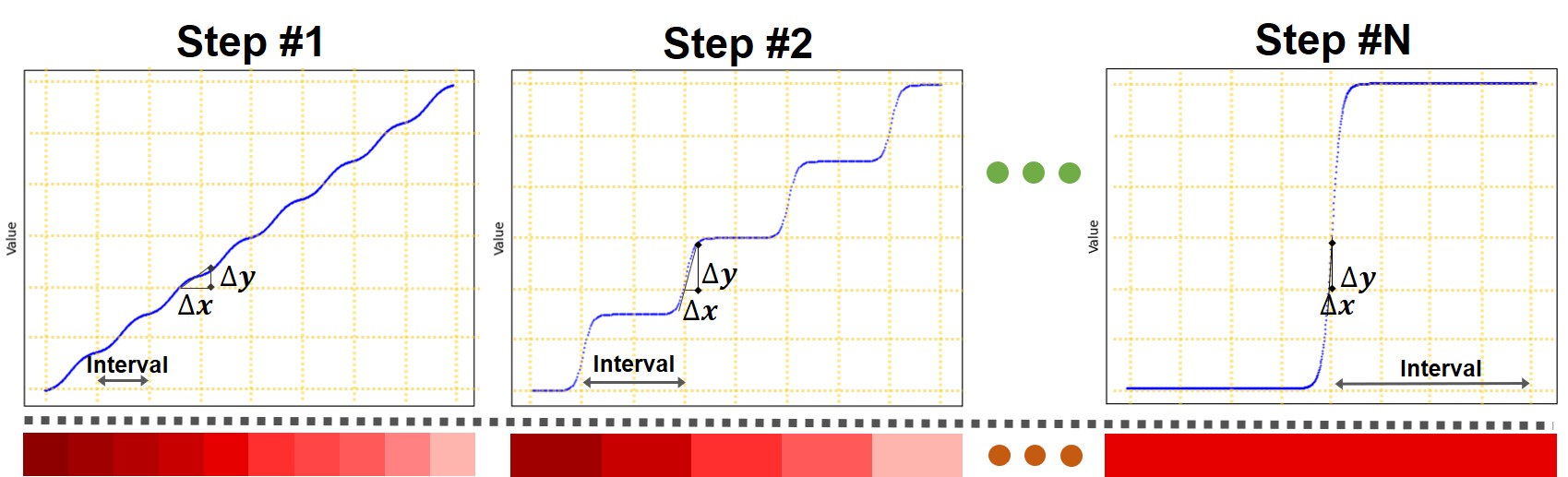}
  \end{center}
  \caption{\textbf{Differential quantization process and updation of quantization table \textit{q}.}  }
  \label{fig:diff-quan}
\end{figure}

In our case, we use a trainable quantization table \textit{q} to quantize the input image $x$ after transformed to frequency domain. Note the purpose of quantization table \textit{q} is the same as $\Delta$. We increase the amount of dropped details by enlarging the interval of quantization table \textit{q}. In order to adjust the quantization table \textit{q} accurately, and further improve the success rate of the proposed attack, we formulate the whole process as an optimization problem by leveraging the gradients of target model $f$ via backward-propagation. However, $\floor{\cdot}$ is a staircase and thus non-differential function, which cannot be integrated to the optimization via back-propagation directly. To tackle this challenge, inspired by Gong. et al.'s \cite{gong2019differentiable} work, we propose a differential asymptotic quantization $\cQ_{diff}(\cdot)$ by introducing tangent function into quantization process to approximate the staircase quantization function gradually, such that $\cQ_{diff}(\cdot) \approx \cQ{(\cdot)}$. Formally, $\cQ_{diff}$ is defined as follows:
\begin{equation}\label{eq:diff_quan}
\cQ_{diff}(x, q) = (\varphi(\frac{x}{q}) + \floor{\frac{x}{q}}) \cdot \textit{q},
\end{equation}
where $\varphi(\cdot)$ approximates the change between two adjacent quantized values. $\varphi(\cdot)$ is continuously differentiable everywhere and defined as follows:
\begin{equation}\label{eq:round2}
\scalebox{0.9}{$
\varphi(\frac{x}{q}) = \frac{1}{2}(1+\tanh{((\frac{x}{q} - \floor{\frac{x}{q} + 0.5})\cdot \log(\frac{2}{\alpha} - 1))} \cdot \log(\frac{1}{1-\alpha})$}),
\end{equation}
where $\alpha$ is an adjustable parameter which controls the steepness of slope between two adjacent quantized values. We decrease the value of $\alpha$ linearly to approximate the staircase function gradually during the optimization (Figure \ref{fig:phi-x}).
The value of $\alpha$ is determined at start and decreased gradually during the optimization, and thus $\cQ_{diff}(\cdot)$ behaves almost the same as the desired staircase quantization function at the end of optimization. Due to page limit, more details regarding $\alpha$ can be referred to work \cite{gong2019differentiable}. Besides, as the quantization table \textit{q} should be integers during the optimization, we update the quantization table \textit{q} with the \textit{sign} of gradients returned via backward propagation. Formally:
\begin{equation}\label{eq:update_q}
\textit{q'} =  \textit{q} + sign(\nabla_q \cL_{adv}(x', y)),\:\: 
\textrm{s.t. }\: \left\|q - q_{init}\right\|_{\infty} <\epsilon
\end{equation}
where the purpose of $\epsilon$ is similar to $l_{p}$-norm, aiming to make the resultant adversarial image $x'$ looking indistinguishable from clean image $x$. In detail, $\epsilon$ confines the norm of quantization table \textit{q}, to further restrict the amount of information to drop. We will give more study about the setting of $\epsilon$ in Section \ref{sec:experiments}.
The illustration of differential quantization and updation of \textit{q} at different steps are shown in Figure \ref{fig:diff-quan}.

During the process of optimization, the interval of \textit{q} increases gradually as updated by Eq. \ref{eq:update_q}. For illustration, we plot color palettes at the bottom to show how the colors are dropped when the interval increases. In summary, the proposed $AdvDrop$ adopts an asymptotic strategy to drop the information, that the slope ($\frac{\Delta y}{\Delta x}$) becomes steeper gradually, thus $\cQ_{diff}(\cdot) \approx \cQ{(\cdot)}$ finally.

\section{Experimental Evaluation}\label{sec:experiments}
We first outline the experimental setup. Then we evaluate our \emph{AdvDrop} regarding its perceptual and attack performance. Afterwards, we evaluate the performance of \emph{AdvDrop} under defense methods. We then analyze $AdvDrop$ via an ablation study. We finally analyze the dropped information by $AdvDrop$ together with attention of the model.
\subsection{Experimental settings}
\textbf{Dataset and models.~} We randomly selected 2000 correctly classified images from ImageNet \cite{deng2009imagenet} to evaluate proposed attack. We use ResNet50 \cite{he2016deep} as the target model for all the experiments. For evaluating the effectiveness of proposed $AdvDrop$ on adversarial training, we used pretrained adversarial model ResNet50 as defense model \cite{robustness} \footnote{\url{https://github.com/MadryLab/robustness}}.

\textbf{Metrics.}
For all the tests we use attack success rate (succ. rate) (\%) as the metric to evaluate the effectiveness of attacks, which is the proportion of successful attacks among the total number of test images defined as $\frac{1}{N}\sum_{n=1}^{N} [f(x) \neq f(x')]$ in untargeted setting, and $\frac{1}{N}\sum_{n=1}^{N} [f(x)=y_{adv}]$ in targeted setting.
Regarding evaluation on the visual quality on attacks, we use Learned Perceptual Image Patch Similarity ($lpips$) metric \cite{zhang2018unreasonable} as the perceptual metric.

\textbf{Baselines. } For perception study in Section \ref{perceptual}, we compared our proposed \emph{AdvDrop} with one of the most commonly used adversarial attack PGD under both $l_2$ and $l_{\infty}$ settings with different constraints. For evaluation under various defenses, we consider defense methods including: feature squeezing \cite{xu2017feature}, pixel deflection \cite{prakash2018deflecting}, JPEG compression \cite{shin2017jpeg} and adversarial training \cite{madry2017towards}. Regarding adversarial attacks to compare, we select several state-of-the-art attacks under both $l_2$ and $l_{\infty}$ settings, including PGD \cite{madry2017towards}, FGSM \cite{goodfellow2014explaining}, C\&W \cite{carlini2017towards}, DeepFool \cite{moosavi2016deepfool}.

\subsection{Perception study}\label{perceptual}
\begin{figure}[htb]
 \begin{center}
    \includegraphics[width = \linewidth]{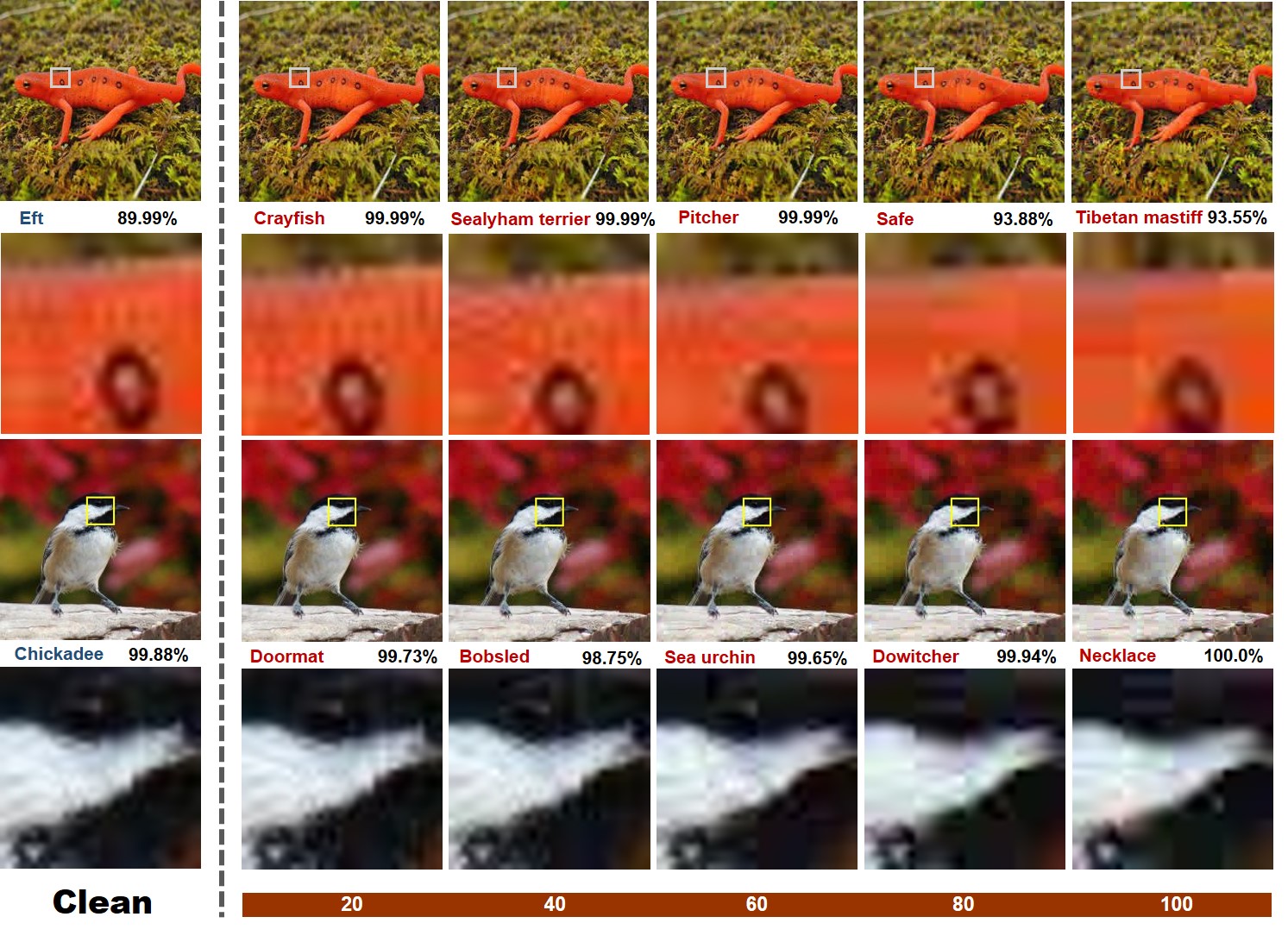}
  \end{center}
  \caption{\textbf{Adv. images generated by \emph{AdvDrop}.} }
  \label{fig:examples-different-q}
\end{figure}

We first conduct perception study on the proposed $AdvDrop$. With enlarging the constraint $\epsilon$ for quantization table \textit{q}, the details disappear gradually as shown in Figure \ref{fig:examples-different-q}.
We then compare the perceptual quality of adversarial examples generated by \emph{AdvDrop} with other attack methods. Though we have a setting (using $\epsilon$) similar to $l_p$ norm, however, our proposed $AdvDrop$ attack optimizes in the frequency domain, that $\epsilon$ is applied as the constraint for \textit{q}. Therefore we adopt $lpips$ \cite{zhang2018unreasonable} as the perceptual metric, which measures how similar the two images are in a way that coincides with human judgment. The value of $lpips$ denotes the perceptual loss, the lower, the better. We compare with one of the most commonly used adversarial attack PGD in both $l_2$ and $l_{\infty}$ settings. 
\begin{figure}[htb]
 \begin{center}
    \includegraphics[width = \linewidth]{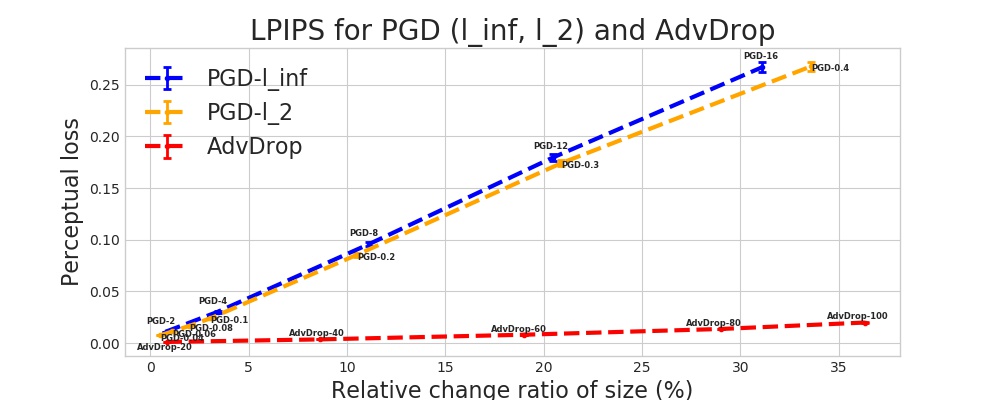}
  \end{center}
  \caption{\textbf{$lpips$ scores for \emph{AdvDrop} and PGD.}}
  \label{fig:lpips}
\end{figure}
    
We summarize the results in Figure \ref{fig:lpips}, where we set \textit{y-axis} with perceptual loss calculated by $lpips$ \cite{zhang2018unreasonable}, and set \textit{x-axis} with the change ratio on size of resultant images compared with clean images. For example, for \emph{AdvDrop}-100, the value of \textit{x-axis} represents the size of adversarial images' decreases by 36.32\% on average compared with clean images' size. Note that opposite to \emph{AdvDrop}, the size of adversarial images generated by PGD is larger than clean images. Thus for PGD, the value on \textit{x-axis} represents how much the ratio in size increases. As Figure \ref{fig:lpips} indicates, though the relative size ratio changes more compared with PGD in either $l_2$ or $l_{\infty}$ settings, the adversarial images generated by $AdvDrop$ are more perceptually aligned with clean images compared with PGD. 

\subsection{Evaluation of \emph{\textbf{AdvDrop}}}
We now evaluate the performance of \emph{AdvDrop} with both targeted and untargeted settings. We evaluate \emph{AdvDrop} with constraint $\epsilon$ for quantization table \textit{q} with 20, 60, 100 respectively. We summarize the results in Table \ref{tab-attack}. 
\begin{table}[htb]
\caption{\textbf{Succ. rate (\%) of $AdvDrop$ on targeted and untargeted settings with different $\epsilon$.}}
\label{tab-attack}
\begin{center}
  \resizebox{\linewidth}{6mm}{ 
\begin{tabular}{l|ccc}
\hline
\textbf{$\epsilon$ for \textit{q}} & \textbf{20} & \textbf{60}  & \textbf{100}   \\ \hline
\textbf{Targeted succ. rate (\%)}       & 97.20 $\pm$ 0.37 & 99.45 $\pm$ 0.16  & 99.95 $\pm$ 0.05 \\ 
\textbf{Untargeted succ. rate (\%)}    & 98.55 $\pm$ 0.26  & 99.85 $\pm$ 0.08 & 100.00 $\pm$ 0.00\\  \hline
\end{tabular}}   
\end{center}
\end{table}

\begin{figure}[htb]
\centering
\begin{subfigure}[b]{0.49\linewidth}
    \includegraphics[width = \linewidth]{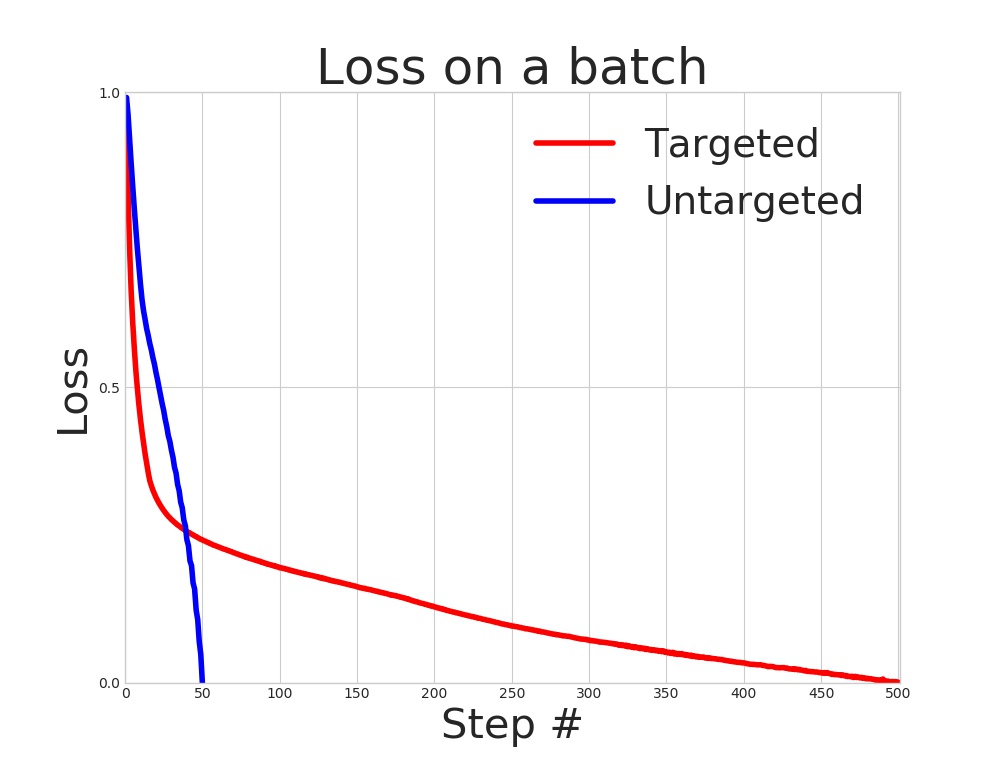}
  \end{subfigure}
\begin{subfigure}[b]{0.49\linewidth}
    \includegraphics[width = \linewidth]{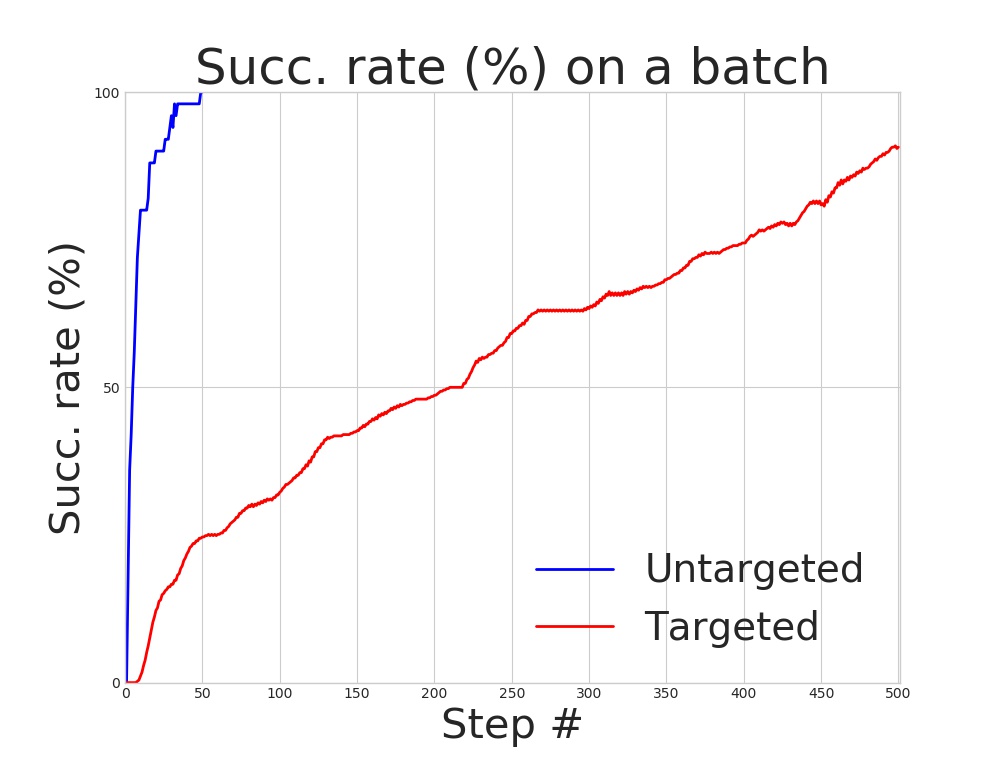}
  \end{subfigure}
  \caption{\textbf{Loss and succ. rate (\%) on a batch.}  }
\label{fig:batch_test}
\end{figure}

As Table \ref{tab-attack} indicates, with relaxing the constraint $\epsilon$, the success rates of \emph{AdvDrop} on both targeted and untargeted settings increase. \emph{AdvDrop} can achieve almost 100\% success rate when $\epsilon=100$. We find that $AdvDrop$ in targeted setting requires more steps to achieve successful attacks compared with untargeted setting (Figure \ref{fig:batch_test}). This may be due to targeted attack always requires more accurate approximation on gradients. During the attack, we set the steps for \emph{AdvDrop} in targeted and untargeted settings with 500 and 50 respectively. We plot the success rates and the loss of a batch on different steps in Figure \ref{fig:batch_test}.

As shown in Figure \ref{fig:batch_test}, the loss of untargeted attack converges rapidly and achieves 100\% success rate when step is around 50. On the other hand, the loss for targeted attack converges until the step is around 500. During the experiments, we find that with the increase of constraint $\epsilon$, fewer steps are required for $AdvDrop$ on both targeted and untargeted settings. For example, in a targeted setting, to achieve success rate above 99\%, on average 496 steps are required when $\epsilon=20$, but only 61 steps are required when $\epsilon=100$.

\subsection{Attack effectiveness under defense methods}
In this part, we evaluate the effectiveness of proposed $AdvDrop$ compared with other adversarial attacks under various defense methods. Here we first generate adversarial examples by adversarial attacks including PGD \cite{madry2017towards}, BIM \cite{dong2018boosting}, C\&W \cite{carlini2017towards}, FGSM \cite{goodfellow2014explaining}, and DeepFool \cite{moosavi2016deepfool}. Regarding these attacks, we consider both settings including $l_2$ and $l_\infty$ to generate adversarial examples. Then we test different defense methods including adversarial training (AT) \cite{athalye2018obfuscated, madry2017towards}, feature squeezing \cite{xu2017feature}, JPEG compression \cite{shin2017jpeg}, and pixel deflection (PD) \cite{prakash2018deflecting} against these samples to evaluate the strength of these attacks under defenses. Among all these defenses, adversarial training is the most effective defense against adversarial attacks. As adversarial training requires too much computation resource, we adopt a black-box setting to evaluate various attacks on adversarial training. We first generate adversarial examples by various attack methods, then feed them to the adversarial trained model. We set $\epsilon=4$ for attacks on $l_{\infty}$ setting, set $\epsilon=0.06$ for attacks on $l_2$ setting that are common used in previous defense methods \cite{prakash2018deflecting,guo2018countering}. We set quantization table \textit{q} with 100 for $AdvDrop$. We summarize the results in Table \ref{tab-defense}.
\begin{table}[htb]
\caption{\textbf{Succ. rate (\%) of attacks under defenses.}}
\label{tab-defense}
 \resizebox{\linewidth}{20mm}{ 
\begin{tabular}{lcccccc}
\hline
\multicolumn{1}{l|}{\multirow{2}{*}{\textbf{Attacks}}} & \multicolumn{1}{c|}{\multirow{2}{*}{\textbf{No Def.}}} & \multirow{2}{*}{\textbf{AT}} & \multicolumn{2}{c}{\textbf{Feature Squeeze}} & \multirow{2}{*}{\textbf{JPEG-30}} & \multirow{2}{*}{\textbf{PD}}  \\ \cline{4-5}
\multicolumn{1}{l|}{}                                  & \multicolumn{1}{c|}{}                                  &                              & \textbf{MF-3}         & \textbf{Bit-6}       &                  &  \\ \hline
\multicolumn{7}{c}{\textbf{$l_{\infty}$}}  \\ \hline
\multicolumn{1}{l|}{\textbf{PGD}}                      & \multicolumn{1}{c|}{100.00}                            & 41.60                        & 90.65                 & 70.50                & 62.50            & 85.10                                \\
\multicolumn{1}{l|}{\textbf{BIM}}                      & \multicolumn{1}{c|}{100.00}                            & 42.80                        & 90.25                 & 69.20                & 33.80            & 81.50                                       \\
\multicolumn{1}{l|}{\textbf{FGSM}}                     & \multicolumn{1}{c|}{91.90}                             & 42.60                        & 91.80                 & 66.05                & 49.60            & 81.40                                      \\ \hline
\multicolumn{7}{c}{\textbf{$l_2$}} \\ \hline
\multicolumn{1}{l|}{\textbf{PGD}}                      & \multicolumn{1}{c|}{100.0}                             & 43.00                        & 90.3                  & 62.8                 & 27.6             & 64.1                                      \\
\multicolumn{1}{l|}{\textbf{Dfool}}                    & \multicolumn{1}{c|}{99.00}                             & 42.95                        & 89.7                  & 29.00                & 21.20            & 12.60                                      \\
\multicolumn{1}{l|}{\textbf{CW}}                       & \multicolumn{1}{c|}{84.30}                             & 43.00                        & 88.80                 & 26.00                & 21.70            & 12.60                                       \\ \hline
\multicolumn{1}{l|}{\textbf{AdvDrop}}                  & \multicolumn{1}{c|}{100.00}                            & \textbf{44.50}               & \textbf{95.35}        & \textbf{82.60}       & \textbf{80.00}   & \textbf{95.65}                        \\ \hline
\end{tabular}}
\end{table}

As Table \ref{tab-defense} shows, since adversarial images are crafted by \emph{AdvDrop} with a totally different paradigm, they are more robust to current defense methods compared to adversarial images generated by other attacks. Among all these defenses, adversarial training is still the most effective defense against \emph{AdvDrop}. We suggest that as the adversarial training makes the model learn more robust feature representations, it is also effective to defend against our proposed $AdvDrop$ to some degree. On the other hand, other denoising-based defense methods demonstrate limited effectiveness to defend against our proposed \emph{AdvDrop}. Though denoising-based strategies show effectiveness in mitigating the adversarial perturbation generated by previous attacks, regarding $AdvDrop$ which has already removed some essential features from clean images, denoising may aggravate the distortion caused by loss of features. We further perform an evaluation by JPEG compression to validate the robustness of adversarial examples generated by \emph{AdvDrop} under denoising operation.

\begin{figure}[htb]
 \begin{center}
    \includegraphics[width =\linewidth]{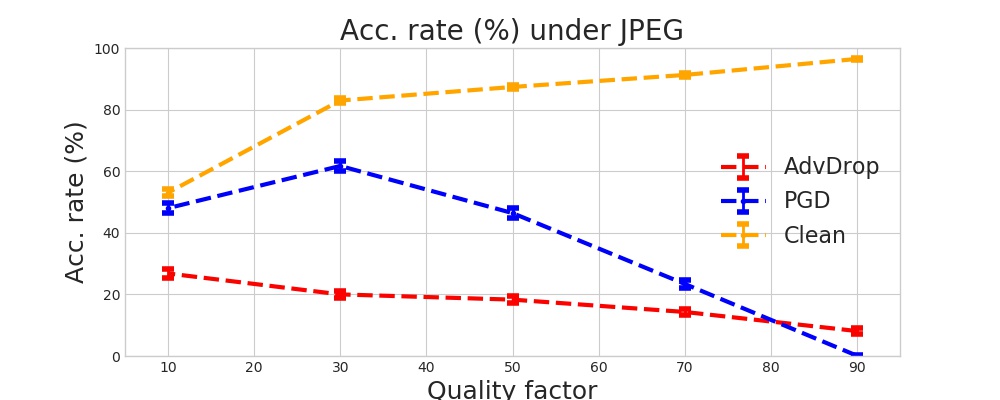}
  \end{center}
  \caption{\textbf{Performance of \emph{AdvDrop} under JPEG.}}
  \label{fig:jpeg}
\end{figure}

\textbf{JPEG compression. } 
Previous studies show that adversarial perturbation can be partly removed via JPEG compression \cite{guo2018countering, das2018shield}. Regarding JPEG compression, the compression rate is controlled by a quantifiable quality, which affects to what extent the information is reduced. 
We evaluate the performance of \emph{AdvDrop} under JPEG compression with different quality factors to represent how robust the adversaries generated by \emph{AdvDrop} are. For comparison, we also test the performance of PGD and the accuracy of clean images under JPEG compression (Figure \ref{fig:jpeg}). We set both \emph{AdvDrop} and PGD with untargeted setting. We report accuracy rate as the metric in Figure \ref{fig:jpeg}, meaning the proportion of recovered adversarial examples by JPEG compression. 

As Figure \ref{fig:jpeg} indicates, when quality factor is extreme low (\eg quality factor = 10), JPEG compression results in corruption on clean images, that clean accuracy even drops below 60\%.
Regarding the adversarial examples generated by PGD, they are mostly recovered when JPEG compression with quality factor being equal to 30. When quality factor further decreases, the recovered rate of PGD decreases due to the corruption caused by JPEG compression. 
Compared to PGD, adversarial examples generated by \emph{AdvDrop} is much less affected by JPEG compression. Less than 30\% adversaries generated by \emph{AdvDrop} are recovered at most. 
We suggest that the mechanism of \emph{AdvDrop} itself is a kind of lossy operation which makes resultant adversarial images lack some key features for recognition. Thus when further applying lossy operations (such as JPEG compression) on resultant adversarial images, the applied lossy operations may even further destroy the generated adversarial examples and even harder to recover. However, there is still chance that the adversarial examples generated by \emph{AdvDrop} can be recovered by ``lossy operation" as Figure \ref{fig:jpeg} shows. 

\subsection{Ablation study}
Here, we conduct experiments to analyze the impact from following aspects for proposed $AdvDrop$ attack: 1) quantization methods, 2) spatial domain, 3) frequency (low frequency, middle frequency or high frequency).

\textbf{Effect of quantization methods.~} 
Here we evaluate how different quantization methods affect the performance of \emph{AdvDrop}. We evaluate a typical rounding method and another differential quantization method proposed by Shin et al. \cite{shin2017jpeg}. During evaluation, we only change the quantization method but keep others of \emph{AdvDrop} unchanged. Typical rounding method $\floor*{x+0.5}$ achieves success rate of only 5.00 $\pm$ 0.98\%. Differential rounding method proposed by Shin et al. \cite{shin2017jpeg}, $\floor*{x+0.5}$ + $(\floor*{x+0.5} - x)^3$ achieves success rate of 65.20 $\pm$ 2.13\%, and ours achieves 97.20 $\pm$ 0.37\% success rate. This demonstrates the effectiveness of differential quantization method adopted by \emph{AdvDrop}.

\textbf{Effect of frequency domain.~}
Here we perform an ablation study to show the advantage of dropping information in the frequency domain rather than spatial domain. We reduce the color depth to drop images' information. We report the accuracy on the same dataset when reducing the image color by different amounts of bits (Table \ref{color-quan}).
\begin{table}[htb]
\caption{\textbf{Acc. rate (\%) by reducing color depth.}}
\label{color-quan}
\begin{center}
 \resizebox{\linewidth}{5mm}{ 
\begin{tabular}{l|ccc}
\hline
\textbf{Bit Depth} & \textbf{2} & \textbf{4}  & \textbf{6}   \\ \hline
\textbf{Acc. rate (\%)} & 11.10 $\pm$ 0.07 & 93.50 $\pm$ 0.05  & 99.90 $\pm$ 0.07 \\ 
\hline
\end{tabular} } 
\end{center}
\end{table}
Compared to dropping information in frequency domain, the accuracy rate is affected until the bit is reduced to 2 (11.10 $\pm$ 0.07\%). As the images' details could be well quantized in the frequency domain, the resultant images are also more natural for human observers.

\textbf{Effect of different frequencies.~}
We also perform an ablation study on how dropping different regions of frequency (low, middle, high) affects the model's clean accuracy. We perform this study by dropping different regions of frequency of given images after transformed to frequency domain by DCT. Results are summarized in Table \ref{tab-frequency}. We also visualize the resultant images and their local details at the bottom of Table \ref{tab-frequency}.
As the results in Table \ref{tab-frequency} indicate, compared to middle and high frequencies, the low frequency part serves as dominant feature for the model. When low frequency is dropped from the images, the accuracy rate drops to 30.00 $\pm$ 1.02\%, but 84.40 $\pm$ 8.11\% and 86.50 $\pm$ 0.76\% when middle and high frequencies dropped. As the images in Table \ref{tab-frequency} show, when low frequency is dropped, the details are almost lost.
\begin{table}[htb]
\caption{\textbf{Acc. rate (\%) by dropping/reserving different frequencies.}}
\label{tab-frequency}
\begin{center}
 \resizebox{\linewidth}{4mm}{ 
\begin{tabular}{l|cccc}
\hline
\textbf{Dropped Freq.} & \textbf{None} &\textbf{Low} & \textbf{Middle} & \textbf{High}   \\ \hline
\textbf{Acc. rate (\%)}   &100.00 $\pm$ 0.00 & 30.00 $\pm$ 1.02  & 84.40 $\pm$ 8.11  & 86.50 $\pm$ 0.76\\ \hline
\end{tabular}}    
\end{center}
\begin{center}
 \includegraphics[width = \linewidth]{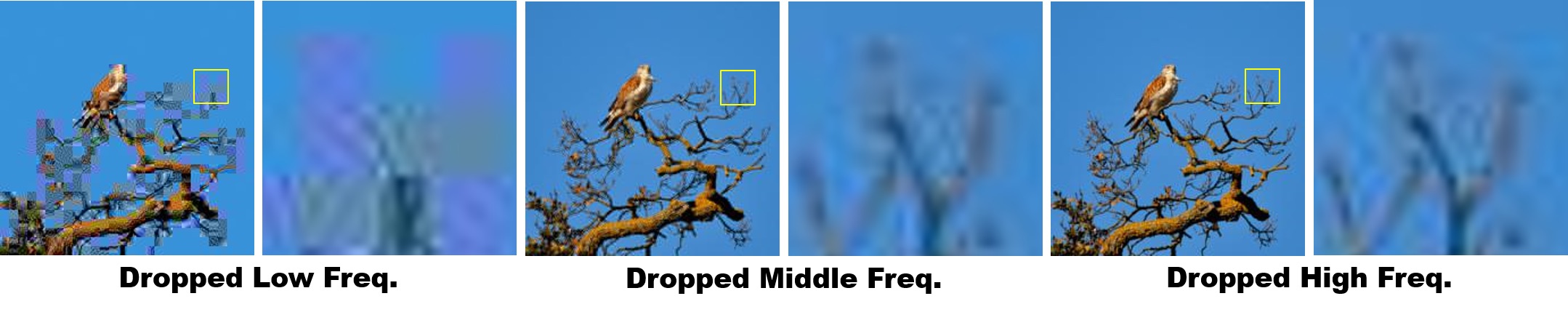}   
\end{center}
\end{table}
\subsection{Visualization and analysis} 
\begin{figure}[htb]
 \begin{center}
    \includegraphics[width =  \linewidth]{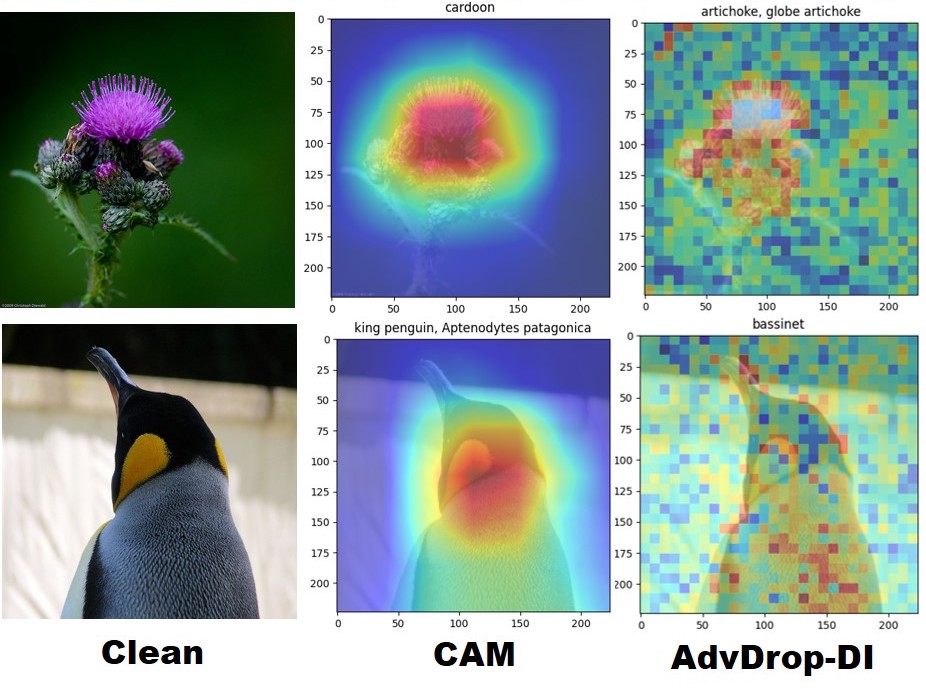}
  \end{center}
  \caption{\textbf{Analysis on dropped information.}  }
  \label{fig:advdrop-di}
\end{figure}

We are also interested in where and what information would be dropped by $AdvDrop$ with a given image? Whether $AdvDrop$ tends to drop the information where models pay attention? Towards this end, we visualize the attention of the model (Grad-CAM \cite{selvaraju2017grad}) and the amount of dropped information by \emph{AdvDrop} on different regions of given images (Figure \ref{fig:advdrop-di}).
Regarding the first case in Figure \ref{fig:advdrop-di}, the model mainly pays attention to the flower part of the ``cardoon", \emph{AdvDrop} drops both calyx and flower parts. In the second case, the model pays attention to the head of the ``peguin", however, in this case, \emph{AdvDrop} mainly throws away the information on the body part of the ``peguin", which has rich texture details regarding the fur of the ``peguin". In summary, there is some overlapping between model's attention and where \emph{AdvDrop} drops information from. A small difference is that \emph{AdvDrop} seems focusing more on the part which has rich texture details. 

We also analyze the components of the dropped information. Here we roughly devide the dropped information as ``high frequency" and ``low frequency". We find \emph{AdvDrop} tends to drop high frequency information than low frequency information. 
\section{Conclusion and Future Work} \label{sec:conclusion}
In this paper, we have investigated the adversarial robustness from a novel perspective, and proposed a novel approach called adversarial drop (\emph{AdvDrop}), which leverages differential quantization and adversarial attack techniques, to craft adversarial examples by dropping existing details of images. \emph{AdvDrop} opens a new way for robustness evaluation of DNNs. 
The proposed \emph{AdvDrop} currently still utilizes a relative simple method to drop the information by focusing on frequency domain. We plan to explore other techniques to drop the information from images in our future work. Also, we will explore how to apply \emph{AdvDrop} on other tasks such as interpretability of DNNs. Moreover, effective defense strategies against \emph{AdvDrop} will be another crucial and promising direction.

\section*{Acknowledgement}
This research is partly supported by ARC Discovery Project under Grant DP180100212. This work was supported by Alibaba Group through Alibaba Research Intern Program.
{\small
\bibliographystyle{ieee_fullname}
\bibliography{egbib}

\begin{thebibliography}{10}\itemsep=-1pt

\bibitem{agustsson2017soft}
Eirikur Agustsson, Fabian Mentzer, Michael Tschannen, Lukas Cavigelli, Radu
  Timofte, Luca Benini, and Luc Van~Gool.
\newblock Soft-to-hard vector quantization for end-to-end learning compressible
  representations.
\newblock {\em NeurIPS}, 2017.

\bibitem{agustsson2019generative}
Eirikur Agustsson, Michael Tschannen, Fabian Mentzer, Radu Timofte, and Luc~Van
  Gool.
\newblock Generative adversarial networks for extreme learned image
  compression.
\newblock In {\em CVPR}, 2019.

\bibitem{ahmed1974discrete}
Nasir Ahmed, T. Natarajan, and Kamisetty~R Rao.
\newblock Discrete cosine transform.
\newblock {\em IEEE Transactions on Computers}, 1974.

\bibitem{athalye2018obfuscated}
Anish Athalye, Nicholas Carlini, and David Wagner.
\newblock Obfuscated gradients give a false sense of security: Circumventing
  defenses to adversarial examples.
\newblock In {\em ICLR}, 2018.

\bibitem{boutell1997png}
Thomas Boutell and T Lane.
\newblock Png (portable network graphics) specification version 1.0.
\newblock {\em Network Working Group}, pages 1--102, 1997.

\bibitem{carlini2017adversarial}
Nicholas Carlini and David Wagner.
\newblock Adversarial examples are not easily detected: Bypassing ten detection
  methods.
\newblock In {\em ACM Workshop on Artificial Intelligence and Security}, pages
  3--14. ACM, 2017.

\bibitem{carlini2017towards}
Nicholas Carlini and David Wagner.
\newblock Towards evaluating the robustness of neural networks.
\newblock In {\em IEEE S\&P}, 2017.

\bibitem{chen2018ead}
Pin-Yu Chen, Yash Sharma, Huan Zhang, Jinfeng Yi, and Cho-Jui Hsieh.
\newblock Ead: elastic-net attacks to deep neural networks via adversarial
  examples.
\newblock In {\em AAAI}, 2018.

\bibitem{dabouei2020smoothfool}
Ali Dabouei, Sobhan Soleymani, Fariborz Taherkhani, Jeremy Dawson, and Nasser
  Nasrabadi.
\newblock Smoothfool: An efficient framework for computing smooth adversarial
  perturbations.
\newblock In {\em Proceedings of the IEEE/CVF Winter Conference on Applications
  of Computer Vision}, pages 2665--2674, 2020.

\bibitem{das2017keeping}
Nilaksh Das, Madhuri Shanbhogue, Shang-Tse Chen, Fred Hohman, Li Chen,
  Michael~E Kounavis, and Duen~Horng Chau.
\newblock Keeping the bad guys out: Protecting and vaccinating deep learning
  with jpeg compression.
\newblock {\em arXiv preprint arXiv:1705.02900}, 2017.

\bibitem{das2018shield}
Nilaksh Das, Madhuri Shanbhogue, Shang-Tse Chen, Fred Hohman, Siwei Li, Li
  Chen, Michael~E Kounavis, and Duen~Horng Chau.
\newblock Shield: Fast, practical defense and vaccination for deep learning
  using jpeg compression.
\newblock In {\em ACM SIGKDD}, 2018.

\bibitem{deng2009imagenet}
Jia Deng, Wei Dong, Richard Socher, Li-Jia Li, Kai Li, and Li Fei-Fei.
\newblock Imagenet: A large-scale hierarchical image database.
\newblock In {\em CVPR}, pages 248--255, 2009.

\bibitem{dong2018boosting}
Yinpeng Dong, Fangzhou Liao, Tianyu Pang, Hang Su, Jun Zhu, Xiaolin Hu, and
  Jianguo Li.
\newblock Boosting adversarial attacks with momentum.
\newblock In {\em CVPR}, 2018.

\bibitem{duan2020adversarial}
Ranjie Duan, Xingjun Ma, Yisen Wang, James Bailey, A.~K. Qin, and Yun Yang.
\newblock Adversarial camouflage: Hiding physical-world attacks with natural
  styles.
\newblock {\em In CVPR}, 2020.

\bibitem{duan2021adversarial}
Ranjie Duan, Xiaofeng Mao, A~Kai Qin, Yuefeng Chen, Shaokai Ye, Yuan He, and
  Yun Yang.
\newblock Adversarial laser beam: Effective physical-world attack to dnns in a
  blink.
\newblock In {\em Proceedings of the IEEE/CVF Conference on Computer Vision and
  Pattern Recognition}, pages 16062--16071, 2021.

\bibitem{dziugaite2016study}
Gintare~Karolina Dziugaite, Zoubin Ghahramani, and Daniel~M Roy.
\newblock A study of the effect of jpg compression on adversarial images.
\newblock {\em arXiv preprint arXiv:1608.00853}, 2016.

\bibitem{robustness}
Logan Engstrom, Andrew Ilyas, Hadi Salman, Shibani Santurkar, and Dimitris
  Tsipras.
\newblock Robustness (python library), 2019.

\bibitem{eykholt2018robust}
Kevin Eykholt, Ivan Evtimov, Earlence Fernandes, Bo Li, Amir Rahmati, Chaowei
  Xiao, Atul Prakash, Tadayoshi Kohno, and Dawn Song.
\newblock Robust physical-world attacks on deep learning visual classification.
\newblock In {\em CVPR}, 2018.

\bibitem{gong2019differentiable}
Ruihao Gong, Xianglong Liu, Shenghu Jiang, Tianxiang Li, Peng Hu, Jiazhen Lin,
  Fengwei Yu, and Junjie Yan.
\newblock Differentiable soft quantization: Bridging full-precision and low-bit
  neural networks.
\newblock In {\em CVPR}, 2019.

\bibitem{goodfellow2014explaining}
Ian~J Goodfellow, Jonathon Shlens, and Christian Szegedy.
\newblock Explaining and harnessing adversarial examples.
\newblock In {\em ICLR}, 2014.

\bibitem{guo2020low}
Chuan Guo, Jared~S Frank, and Kilian~Q Weinberger.
\newblock Low frequency adversarial perturbation.
\newblock In {\em Uncertainty in Artificial Intelligence}, 2020.

\bibitem{guo2018countering}
Chuan Guo, Mayank Rana, Moustapha Cisse, and Laurens van~der Maaten.
\newblock Countering adversarial images using input transformations.
\newblock In {\em ICLR}, 2018.

\bibitem{guo2020watch}
Qing Guo, Felix Juefei-Xu, Xiaofei Xie, Lei Ma, Jian Wang, Bing Yu, Wei Feng,
  and Yang Liu.
\newblock Watch out! motion is blurring the vision of your deep neural
  networks.
\newblock {\em In NeurIPS}, 33, 2020.

\bibitem{he2016deep}
Kaiming He, Xiangyu Zhang, Shaoqing Ren, and Jian Sun.
\newblock Deep residual learning for image recognition.
\newblock In {\em CVPR}, 2016.

\bibitem{heckbert1982color}
Paul Heckbert.
\newblock Color image quantization for frame buffer display.
\newblock {\em ACM Siggraph Computer Graphics}, pages 297--307, 1982.

\bibitem{hosseini2018semantic}
Hossein Hosseini and Radha Poovendran.
\newblock Semantic adversarial examples.
\newblock In {\em CVPR Workshop}, 2018.

\bibitem{ilyas2019adversarial}
Andrew Ilyas, Shibani Santurkar, Logan Engstrom, Brandon Tran, and Aleksander
  Madry.
\newblock Adversarial examples are not bugs, they are features.
\newblock {\em In NeurIPS}, 2019.

\bibitem{johnston2018improved}
Nick Johnston, Damien Vincent, David Minnen, Michele Covell, Saurabh Singh,
  Troy Chinen, Sung~Jin Hwang, Joel Shor, and George Toderici.
\newblock Improved lossy image compression with priming and spatially adaptive
  bit rates for recurrent networks.
\newblock In {\em CVPR}, 2018.

\bibitem{kurakin2016adversarial}
Alexey Kurakin, Ian Goodfellow, and Samy Bengio.
\newblock Adversarial examples in the physical world.
\newblock {\em ICLR}, 2016.

\bibitem{li2018learning}
Mu Li, Wangmeng Zuo, Shuhang Gu, Debin Zhao, and David Zhang.
\newblock Learning convolutional networks for content-weighted image
  compression.
\newblock In {\em CVPR}, 2018.

\bibitem{liu2018beyond}
Hsueh-Ti~Derek Liu, Michael Tao, Chun-Liang Li, Derek Nowrouzezahrai, and Alec
  Jacobson.
\newblock Beyond pixel norm-balls: Parametric adversaries using an analytically
  differentiable renderer.
\newblock In {\em ICLR}, 2018.

\bibitem{madry2017towards}
Aleksander Madry, Aleksandar Makelov, Ludwig Schmidt, Dimitris Tsipras, and
  Adrian Vladu.
\newblock Towards deep learning models resistant to adversarial attacks.
\newblock In {\em ICLR}, 2018.

\bibitem{moosavi2016deepfool}
Seyed-Mohsen Moosavi-Dezfooli, Alhussein Fawzi, and Pascal Frossard.
\newblock Deepfool: a simple and accurate method to fool deep neural networks.
\newblock In {\em CVPR}, 2016.

\bibitem{orchard1991color}
Michael~T Orchard, Charles~A Bouman, et~al.
\newblock Color quantization of images.
\newblock {\em IEEE Transactions on Signal Processing}, pages 2677--2690, 1991.

\bibitem{prakash2018deflecting}
Aaditya Prakash, Nick Moran, Solomon Garber, Antonella DiLillo, and James
  Storer.
\newblock Deflecting adversarial attacks with pixel deflection.
\newblock In {\em CVPR}, 2018.

\bibitem{selvaraju2017grad}
Ramprasaath~R Selvaraju, Michael Cogswell, Abhishek Das, Ramakrishna Vedantam,
  Devi Parikh, and Dhruv Batra.
\newblock Grad-cam: Visual explanations from deep networks via gradient-based
  localization.
\newblock In {\em ICCV}, pages 618--626, 2017.

\bibitem{shamsabadi2020colorfool}
Ali~Shahin Shamsabadi, Ricardo Sanchez-Matilla, and Andrea Cavallaro.
\newblock Colorfool: Semantic adversarial colorization.
\newblock In {\em CVPR}, 2020.

\bibitem{sharif2016accessorize}
Mahmood Sharif, Sruti Bhagavatula, Lujo Bauer, and Michael~K Reiter.
\newblock Accessorize to a crime: Real and stealthy attacks on state-of-the-art
  face recognition.
\newblock In {\em CCS}, 2016.

\bibitem{sharma2019effectiveness}
Yash Sharma, Gavin~Weiguang Ding, and Marcus Brubaker.
\newblock On the effectiveness of low frequency perturbations.
\newblock {\em arXiv preprint arXiv:1903.00073}, 2019.

\bibitem{shin2017jpeg}
Richard Shin and Dawn Song.
\newblock Jpeg-resistant adversarial images.
\newblock In {\em NeurIPS Workshop}, 2017.

\bibitem{skodras2001jpeg}
Athanassios Skodras, Charilaos Christopoulos, and Touradj Ebrahimi.
\newblock The jpeg 2000 still image compression standard.
\newblock {\em IEEE Signal Processing Magazine}, 2001.

\bibitem{szegedy2013intriguing}
Christian Szegedy, Wojciech Zaremba, Ilya Sutskever, Joan Bruna, Dumitru Erhan,
  Ian Goodfellow, and Rob Fergus.
\newblock Intriguing properties of neural networks.
\newblock In {\em ICLR}, 2013.

\bibitem{toderici2017full}
George Toderici, Damien Vincent, Nick Johnston, Sung Jin~Hwang, David Minnen,
  Joel Shor, and Michele Covell.
\newblock Full resolution image compression with recurrent neural networks.
\newblock In {\em CVPR}, 2017.

\bibitem{wallace1992jpeg}
Gregory~K Wallace.
\newblock The jpeg still picture compression standard.
\newblock {\em IEEE Transactions on Consumer Electronics}, 1992.

\bibitem{wiyatno2019physical}
Rey~Reza Wiyatno and Anqi Xu.
\newblock Physical adversarial textures that fool visual object tracking.
\newblock In {\em ICCV}, 2019.

\bibitem{xu2017feature}
Weilin Xu, David Evans, and Yanjun Qi.
\newblock Feature squeezing: Detecting adversarial examples in deep neural
  networks.
\newblock {\em NDSS}, 2017.

\bibitem{zeng2019dirichlet}
Min Zeng, Yisen Wang, and Yuan Luo.
\newblock Dirichlet latent variable hierarchical recurrent encoder-decoder in
  dialogue generation.
\newblock In {\em EMNLP}, 2019.

\bibitem{zeng2019adversarial}
Xiaohui Zeng, Chenxi Liu, Yu-Siang Wang, Weichao Qiu, Lingxi Xie, Yu-Wing Tai,
  Chi-Keung Tang, and Alan~L Yuille.
\newblock Adversarial attacks beyond the image space.
\newblock In {\em CVPR}, 2019.

\bibitem{zhang2018unreasonable}
Richard Zhang, Phillip Isola, Alexei~A Efros, Eli Shechtman, and Oliver Wang.
\newblock The unreasonable effectiveness of deep features as a perceptual
  metric.
\newblock In {\em CVPR}, pages 586--595, 2018.

\bibitem{zhao2018generating}
Zhengli Zhao, Dheeru Dua, and Sameer Singh.
\newblock Generating natural adversarial examples.
\newblock In {\em International Conference on Learning Representations}, 2018.

\bibitem{zhao2020towards}
Zhengyu Zhao, Zhuoran Liu, and Martha Larson.
\newblock Towards large yet imperceptible adversarial image perturbations with
  perceptual color distance.
\newblock In {\em CVPR}, 2020.

\end{thebibliography}
}

\end{document}